\documentclass[10pt,twocolumn]{article}

\usepackage[review]{cvpr}      
\usepackage{multirow}
\definecolor{cvprblue}{rgb}{0.21,0.49,0.74}
\usepackage[pagebackref,breaklinks,colorlinks,allcolors=cvprblue]{hyperref}

\title{Automatic Spectral Calibration of Hyperspectral Images:\\ Method, Dataset and Benchmark}

\author{Zhuoran Du \\
Beijing Jiaotong University\\
{\tt\small 20112007@bjtu.edu.cn}
\and
Shaodi You\\
University of Amsterdam\\
{\tt\small s.you@uva.nl}
\and
Cheng Cheng\\
Beijing Jiaotong University\\
{\tt\small 23125171@bjtu.edu.cn}
\and
Shikui Wei\\
Beijing Jiaotong University\\
{\tt\small shkwei@bjtu.edu.cn}
}

\begin{document}
\date{}
\maketitle
\begin{abstract}
Hyperspectral image (HSI) densely samples the world in both the space and frequency domain and therefore is more distinctive than RGB images. 
Usually, HSI needs to be calibrated to minimize the impact of various illumination conditions. 
The traditional way to calibrate HSI utilizes a physical reference, which involves manual operations, occlusions, and/or limits camera mobility.
These limitations inspire this paper to automatically calibrate HSIs using a learning-based method.
Towards this goal, a large-scale HSI calibration dataset is created, which has 765 high-quality HSI pairs covering diversified natural scenes and illuminations. The dataset is further expanded to 7650 pairs by combining with 10 different physically measured illuminations.
A spectral illumination transformer (SIT) together with an illumination attention module is proposed. 
Extensive benchmarks demonstrate the SoTA performance of the proposed SIT. 
The benchmarks also indicate that low-light conditions are more challenging than normal conditions.
The dataset and codes are available online: \url{https://github.com/duranze/Automatic-spectral-calibration-of-HSI}

\end{abstract}    
\section{Introduction}
\label{sec:intro}

\begin{figure}[t]
  \centering
\includegraphics[width=.99\linewidth]{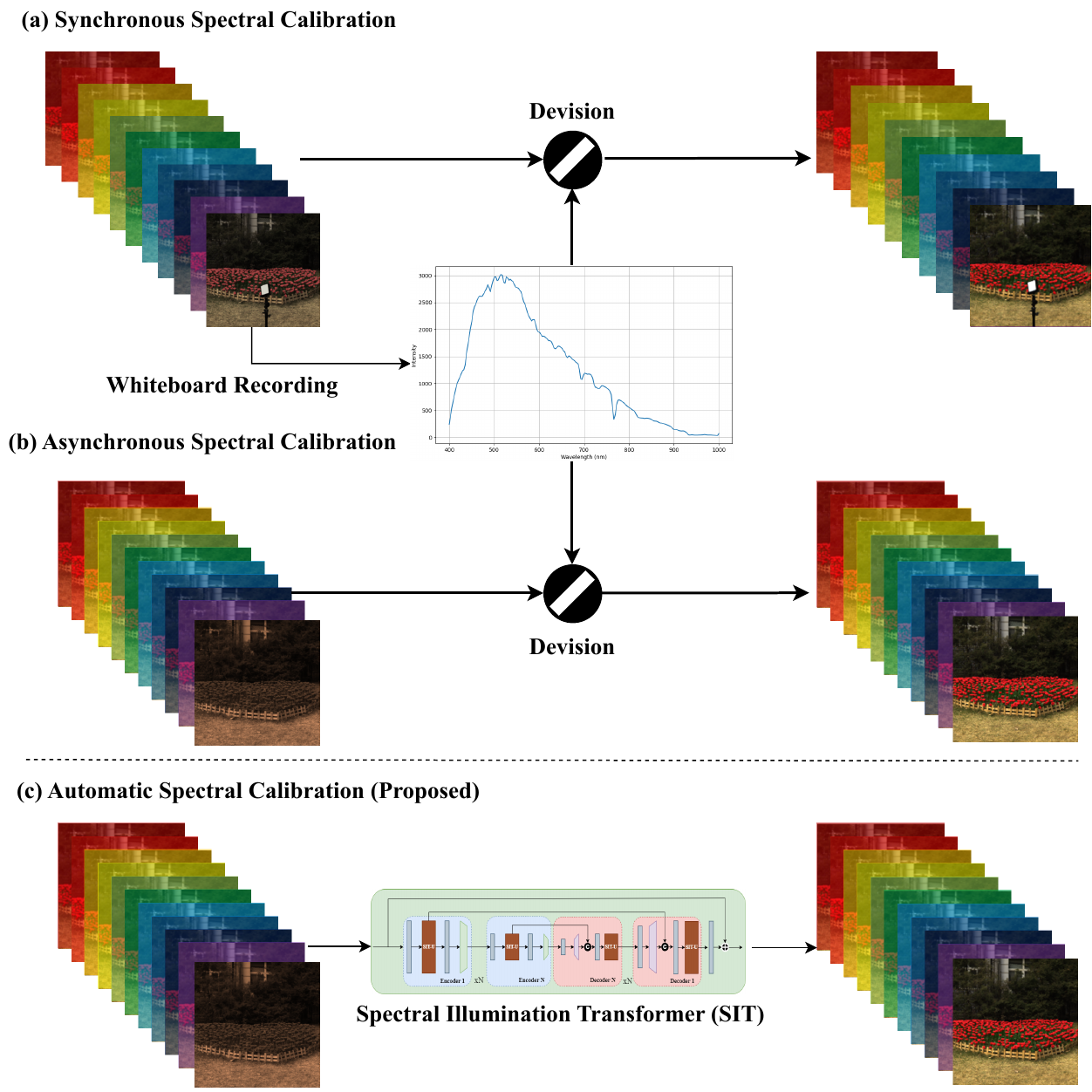}
  \caption{Comparison of HSI Calibration Methods: (a) Synchronous, (b) Asynchronous, and (c) Automatic.}
  \label{fig-scenes}
\end{figure}

Hyperspectral image (HSI) densely samples the world in both spatial and frequency domains. 
It is more informative than RGB images for recognizing material. And therefore it is widely used in remote sensing and laboratory analysis \cite{losurdo2009spectroscopic}\cite{ben2018characterization}\cite{carter2017atomic}\cite{li2021spectral}. 

However, HSI appearance is highly dependent on global illumination. While in remote sensing and lab environments, global illuminations are highly uniform in a large spatial and temporal range or can be well controlled; global illuminations in natural scenes are highly varied spatially and temporally. Therefore, to capture HSI in natural scenes, global illumination needs to be measured simultaneously.

There are two traditional ways to measure the global illumination. The synchronous method \cite{ntire2022} (Fig.~\ref{fig-scenes}.a) utilizes a \ reference such as a barium sulfate panel (an ideal white panel) \cite{leshkevich1988goniometric} in the scene. The disadvantage is the panel partly occludes the scene. The asynchronous method \cite{du2023,icvl} (Fig.~\ref{fig-scenes}.b) takes two HSIs of the same scenes quickly with and without the reference. The disadvantage is that this method requires double shots, manual operations, and an unmoved camera.

These disadvantages inspire this paper: \textit{can we learn to automatically calibrate the HSI without a physical reference?} If the accuracy is close to the physical methods, then the learning-based method has the advantages of less manual operations, no occlusion, and more mobility.

To enable this research, we have created the first HSI auto calibration dataset \textit{BJTU-UVA Dataset}. The dataset is captured using a 204 bands 400-1000nm hyperspectral camera with a 512 $\times$ 512 spatial resolution, and 12bits quantization. The dataset has 765 calibrated and uncalibrated image pairs which are obtained using the asynchronous method. The dataset covers diversified scenes and illuminations. 
To further study the illumination calibration, an expansion dataset \textit{BJTU-UVA-E} is created, which composite calibrated HSIs with 10 different physically measured illuminations. In such a way, the BJTU-UVA-E has 7650 uncalibrated HSIs.
It is worth noticing that BJTU-UVA is not only the first HSI calibration dataset, it is also one of the largest public HSI datasets.

After that, we propose the Spectral Illumination Transformer (SIT), which is the first learning-based HSI calibration method to the best of our knowledge. SIT is based on a U-shape transformer. To allow SIT to learn illumination properly, we propose the \textit{illumination attention} module which is inspired by the classic Gray-World\cite{grayworld} algorithm. We consider SIT the baseline of learning-based methods which may inspire future research in this new topic.

Detailed benchmarks are provided using the BJTU-UVA dataset. Given that there are no other learning-based methods for this novel task, methods for HSI denoising and reconstruction are compared. And classic method of Gray-World is compared. It is shown that SIT is the state-of-the-art. It is interesting to find that Gray-World performs comparable to other learning-based methods on real illuminations. It is also noticed that low-light scenes/areas are likely to have larger errors; which can be a future research topic.

The novelty of the paper can be summarized as:

\begin{itemize}
    \item We propose a novel research topic automatic HSI calibration, which has no occlusion, less manual operations, and more mobility than traditional methods.
    \item We propose the large-scale HSI dataset BJTU-UVA, which is the first dataset for HSI auto-calibration.
    \item We propose the spectral illumination transformer (SIT) with the illumination attention module, which is the very first transformer-based method on this topic.
\end{itemize}




\section{Related Work}
\label{sec:related}

\subsection{Hyperspectral Image Calibration}

In the literature, several spectral calibrations using physical references are developed for remote sensing and indoor optics.

In remote sensing, calibration is done using a physical reference (Fig.~\ref{fig-scenes}.a). Pan et al. in 2001 \cite{remer2001angular} used a spectroradiometer calibrated against a barium sulfate panel (the physical reference), assuming constant solar irradiance; Ferreira et al. in 2004 \cite{ferreira2004optical} used linear interpolation of pre- and post-flight reference panel readings; and Miura et al. in 2008 \cite{miura2008technique} used continuous ground-based panel readings during flight to adjust data.
The limitation of applying this method to natural scenes is that there always exists occlusion from the white reference.

In laboratory settings, where illumination can be precisely controlled, calibration can be performed asynchronously (Fig.~\ref{fig-scenes}.b). Under these conditions, illumination only needs to be measured once and can then be applied in subsequent measurements.
Recently, Moghadam et al. \cite{indoorhsi} proposed a data-driven method to enable indoor hyperspectral imaging using affordable and commonly available LED and fluorescent lighting sources.
However, in the natural outdoor scenes, controlling illumination is infeasible. Therefore, to apply asynchronous methods, after measuring the illumination, the HSI has to be taken as soon as possible. This requires manual operations and a stable camera.


\subsection{RGB White Balancing}

RGB white balancing is a similar topic but with major differences. First, each of the RGB channels is a wide-band channel whereas in HSI the channel is a narrow band. Second, RGB bands usually overlap with each other whereas HSI bands do not or have limited overlaps. Third, many RGB-based methods aim for optimizing human perception while HSI calibration aims for physical accuracy.

Recent advances in white balancing for sRGB images have primarily leveraged deep learning techniques to enhance color correction. 
The two-stage framework by Farghaly et al. \citep{farghaly2023two} combines global color mapping with local adjustments, achieving perceptually improved results. 
Stability across varying color temperatures has been addressed by Li et al. \cite{li2023swbnet} with their SWBNet, which learns color temperature-insensitive features and applies a novel contrastive loss for consistent outcomes. 
Addressing the complexity of mixed-illuminant scenes, Afifi et al. \cite{afifi2022auto} blend images with predefined white-balance settings using learned weighting maps, while Kinli et al. \cite{kinli2023modeling} model lighting variations as style factors, enhancing correction capabilities without explicit illuminant detection.
Lastly, efficiency in processing is significantly enhanced by ~\cite{kinli2023deterministic} through deterministic illumination mapping, facilitating real-time applications. 

\subsection{HSI Restoration and Reconstruction}

Learning-based research on HSI is mainly focusing on denoising and band reconstruction. This research explores many new deep-learning architectures for HSI because it is not trivial to extend RGB-based architectures. Notice that this research does not need the HSI to be calibrated and cannot directly calibrate the HSI.

Puria et al. proposed the DivIll \cite{indoorhsi} method, which uses CNN layers and residual blocks to address reflectance abnormalities caused by illumination variations in indoor scenes.  
Li et al. \cite{li2023hyperspectral} introduces a local-area-based approach that linearly represents each pixel by its neighbors without requiring parameterized weight calculations, emphasizing spectrum preservation. 
Zhang et al. \cite{zhang2023hyperspectral} provide a comprehensive review of HSI denoising techniques ranging from model-driven to hybrid model-data-driven methods. 
Lastly, Pan et al. \cite{pan2023multiscale} integrates multiscale contextual information with a coattention mechanism in the Multiscale Adaptive Fusion Network (MAFNet), showcasing improved denoising capabilities.
SERT \cite{sert} introduces a rectangle self-attention mechanism to capture non-local spatial similarity and a spectral enhancement module to extract global low-rank properties of hyperspectral images for effective noise suppression.
Hu~\etal\cite{hcanet} proposed the HCANet by extending the U-shaped transformer to hyperspectral imaging (HSI) and leveraged a spectral transformer to capture inter-channel relationships effectively.

\section{HSI Calibration}
\label{sec:problem}

In this paper, we denote an HSI as: \protect{$I(x,y,\lambda)\in \mathbb{R}^{H\times W \times C}$}, where $I$ is the intensity, $x, y$ are the spatial coordinates and $\lambda$ the wavelength; $H,W, C$ are the height, width, and channels correspondingly.
In this paper we focus on global illumination and assume it is uniform, therefore, for simplicity, in most of this paper, we write only the wavelength $I(\lambda)$.

\paragraph{HSI Illumination Calibration}

Physically, an HSI $I$ captures the production of scene reflectance $R$ and illumination $L$ wavelength-wisely:
\begin{equation}
\label{eq:phy}
I(\lambda) = R(\lambda) L(\lambda).  
\end{equation}

Unlike RGB, scene reflectance $R(\lambda)$ is highly distinctive if measured in an HSI way. And therefore it is widely used in remote sensing \cite{losurdo2009spectroscopic}\cite{ben2018characterization}\cite{carter2017atomic}\cite{li2021spectral} and laboratory analysis\cite{salas2013new}\cite{elmasry2016image}.
Notice the reflectance is not albedo but still varies from shading and inter-flection and non-uniform illumination.

The goal of HSI calibration is to linearly adjust the intensity as if the global illumination is ideal white in the sampled wavelength range:
\begin{equation}
{L(\lambda) \equiv l},
\end{equation}
where $l$ is a constant which usually set to be $l=1$, then $R(\lambda) \equiv I(\lambda)$.

If the illumination $I(\lambda)$ is not ideal white, then we can calibrate the image as if the illumination is ideal white by:
\begin{equation}
\label{eq:cal_r}
R(\lambda) = \frac{I(\lambda)}{L(\lambda) }.  
\end{equation}

\paragraph{Synchronous method}
In a synchronized method (Fig.~\ref{fig-scenes}.a), $L(\lambda)$ is measured directly using a reference such as a barium sulfate panel \cite{leshkevich1988goniometric}. 
The disadvantage is the reference has to be in the image which causes occultation.

\paragraph{Asynchronous method}
In the asynchronous method, we first measure $L_{p, t_1}(\lambda)$ on the reference at position $p$ and time $t_1$. Then, take an HSI $I_{p, t_2}(\lambda)$ without the reference at the same position but at a later time. Assuming the difference between $L_{p,t_1}(\lambda)$ and $L_{p,t_2}(\lambda)$ is negligible, then: 
$R_2(\lambda) = \frac{I_{p,t_2}(\lambda)}{L_{p,t_1}(\lambda)}$.
The disadvantages are more manual operations; and in an outdoor natural scene, the illumination can be very dynamic, and therefore needs to be remeasured frequently; it also makes moving the camera infeasible.

\paragraph{Learning-based method (proposed)}
Considering the above disadvantages, we propose a novel research question: can the scene radiance be inferred directly through a learning method $f$ without a physical reference and have satisfying accuracy as a physical measurement? 
\begin{equation}
\tilde{R}(\lambda) = f(I(\lambda)).
\label{eq:learning}
\end{equation}
If possible it has at least three advantages: first, there is no occlusion, second camera can move freely, third the imaging only needs to be done once.

\section{BJTU-UVA Dataset}
\label{sec:dataset}


Our work introduces the first-ever dataset for automatic spectral calibration. The  \textit{hyperspectral automatic spectral calibration image} dataset (BJTU-UVA) comprises 765 pairs of real HSI pairs; each pair includes one original HSI and its asynchronous calibration. 
To further diversify the illuminations, we combine the scene radiance with ten different real illuminations to generate more uncalibrated images. The expanded dataset (BJTU-UVA-E) comprises 7650 pairs of uncalibrated and calibrated HSIs.

\subsection{BJTU-UVA Dataset}

For the HSI camera, we use Specim IQ 
\cite{specim_website}, which offers a spectral resolution of 3nm across a wavelength range from 400nm to 1000nm. For reference, we use the IQ White Reference Board \cite{specim_website}, which can reflect all bands of the spectrum from 400nm to 1000nm evenly.

For caption, we use the asynchronous method as described in Sec.~3. 
Specifically, we capture the scene $I_{p,t}(\lambda)$ at position $p$ and time $t$ without the reference. And right after that, we capture the illumination $L_{p,t+\delta}(\lambda)$ by putting the reference in the scene. We kept the time difference small and the camera position unchanged so as to minimize the change of the illumination. The images are stored in the linear format without gamma correct in 12 bits.


In practice, the scene and illumination are not noise-free. The major source of the noise is the dark current $d(\lambda)$. Dark current refers to the electronic noise recorded by the camera's sensor even when no light is present. $d(\lambda)$ is also recorded.


The ground truth is thus generated with the consideration of $d(\lambda)$:
\begin{equation}
\label{eq:cal_rs}
R_{gt}(\lambda) = \frac{I_{p,t}(\lambda)}{L_{p,t+\delta}(\lambda)} = \frac{I^{d}_{p,t}(\lambda)-d_{p,t}(\lambda)}{L^{d}_{p,t+\delta}(\lambda)-d_{p,t+\delta}(\lambda)}.
\end{equation}

\begin{figure*}
  \centering
\includegraphics[width=.99\linewidth]{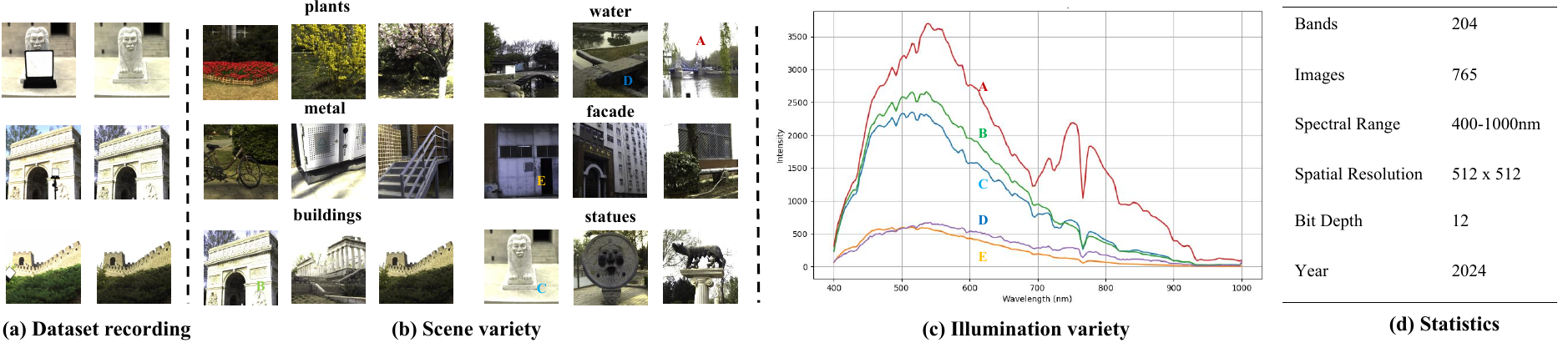}
  \caption{Overview of the BJTU-UVA Dataset: (a) Dataset Recording Setup, (b) Variety of Scenes, (c) Illumination variety, and (d) HSI Camera Specifications and Dataset Statistics}
  \label{fig-dataset}
\end{figure*}

The dataset involves diversified scenes and illuminations as shown in Fig. \ref{fig-dataset}. These scenes feature urban and natural environments under different weather conditions and times of day, including roads, buildings, statues, flowers, and other colored materials. The variety in scene composition ensures that the dataset can adequately test the efficacy of spectral calibration methods.

\begin{figure}
  \centering
\includegraphics[width=.99\linewidth]{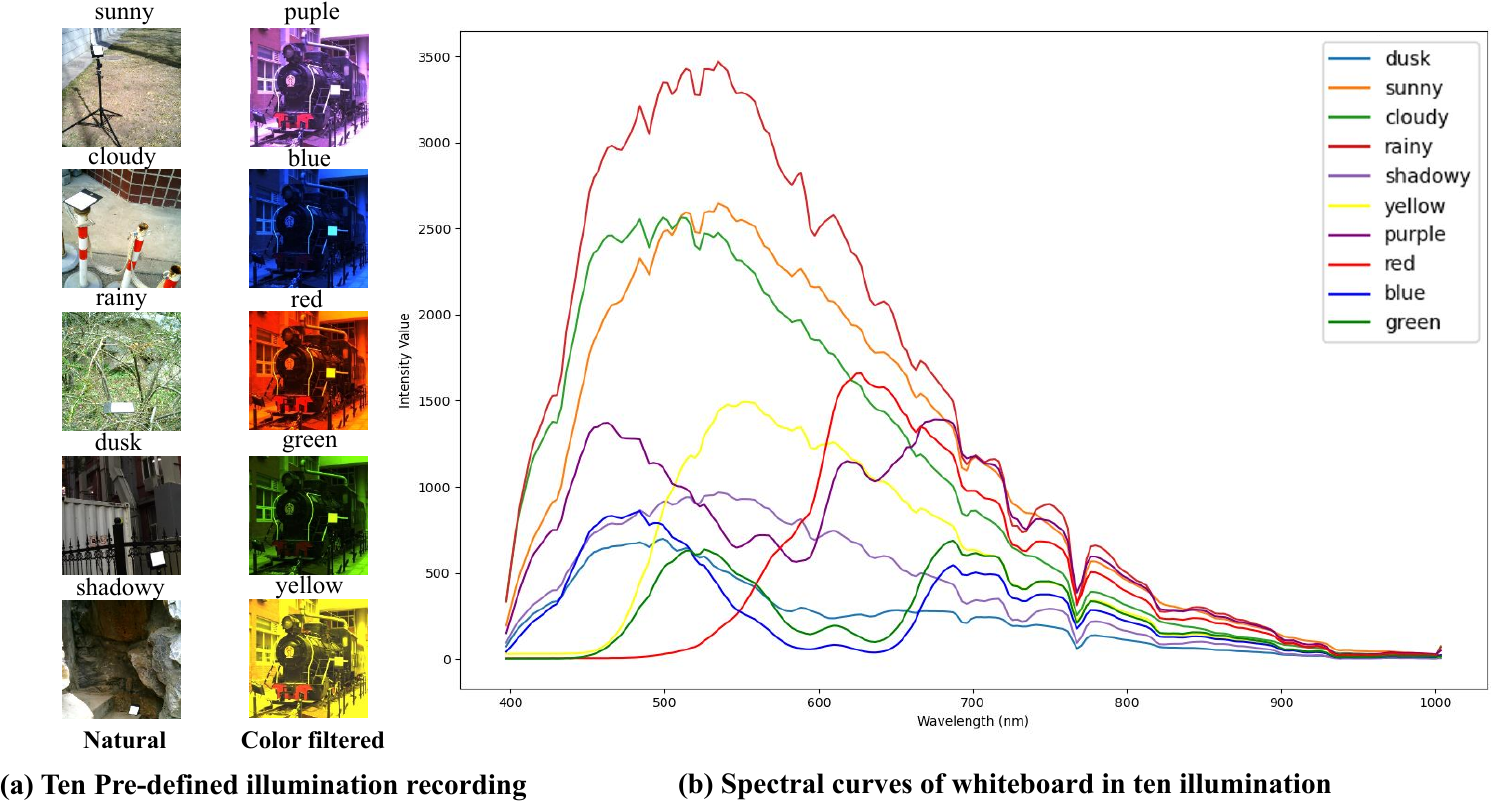}
  \caption{
  Measurement of Ten Different Illuminations Using a Whiteboard: Five Natural Conditions (Sunny, Cloudy, Rainy, Evening, and Shadow) and Five Color-Filtered Conditions (Red, Blue, Yellow, Purple, and Green)
  }
  \label{illu-record}
\end{figure}





\subsection{BJTU-UVA-E Dataset}

To enable further study of the illumination. We expand the dataset by combining the ground truth reflectance $R_{gt}(\lambda)$ with 10 different physically measured illuminations $L_{S_{i}}(\lambda)$.
\begin{equation}
I_{S_{i}}(\lambda) = R_{gt}(\lambda) L_{S_{i}}(\lambda), i \in 1,2, \cdots 10,
\end{equation}

We record these ten different illuminations with the whiteboard as shown in Fig. \ref{illu-record}. To minimize the dark current, we also deduct the noise $d_{S_{i}}(\lambda)$ from actual illuminance recording $L^{d}_{S_{i}}(\lambda)$ by
\begin{equation}
L_{S_{i}}(\lambda) = L^{d}_{S_{i}}(\lambda)-d_{S_{i}}(\lambda),  i \in 1,2, \cdots 10.
\end{equation}


As shown in Fig.~\ref{illu-record}, the 10 illuminations include five natural illuminations: sunny, cloudy, rainy, evening, and shadow; which are measured using the whiteboard in the scenes.
The rest five are colored filtered, which are measured by placing the color filters (red, blue, yellow, purple, and green) in front of the whiteboard.


\subsection{31 Channels Subset}

Since much of the existing HSI research is conducted using 31 channels spanning 400–700 nm, we resample our HSIs to match this standard, covering 31 channels from 400–700 nm, for both the BJTU-UVA and BJTU-UVA-E datasets. This approach ensures compatibility with legacy methods.

\subsection{Specifications}
The dataset specifications are summarized in Fig.~\ref{fig-dataset}.d. Notably, this is not only the first HSI illumination calibration dataset but also one of the largest HSI datasets focused on natural scenes (further details are provided in the supplementary materials).

\subsection{Evaluation Metrics}

We introduce four commonly used metrics: PSNR{\cite{psnr}}, RMSE{\cite{rmse}}, ERGAS{\cite{ergas}}, and SAM{\cite{sam}}. 

The Peak Signal-to-Noise Ratio (PSNR) for HSI is defined as:
\begin{equation}
\mathrm{PSNR}= \frac{1}{B} \sum_{k=1}^B 10 \log _{10}\left(\frac{\max \left(R_{gt}^{k}\right)^2}{\frac{1}{H W}\left\|R_{gt}^{k}-\hat{R}^{k}\right\|_2^2}\right),
\end{equation}
in which $B$ denotes the number of bands and $\left\| \cdot\right\|_2$ refers to the two-norm.

The Root Mean Squared Error (RMSE) is calculated by:
\begin{equation}
\mathrm{RMSE} = \sqrt{\frac{1}{N} \sum_{i=1}^N (\hat{R_{gt}}^{i}-R^{i})^{2}},
\end{equation}
in which $N$ denoted the number of pixels in each image. 

The Error Relative Global Dimensionless Synthesis (ERGAS) metric is specifically designed for evaluating the quality of high-resolution synthesized images and can be calculated by:
\begin{equation}
\mathrm{ERGAS}= \sqrt{\frac{1}{B} \sum_{k=1}^B \frac{\left\|R^{k}_{gt}-\hat{R}^{k}\right\|_2^2}{\mu^2\left(R_{gt}^{k}\right)}},
\end{equation}
in which $\mu$ denotes the mean operation, and $B$ number of bands.

The Spectral Angle Mapper (SAM) is commonly employed to assess the extent of spectral information preservation at each pixel, which can be calculated by:
\begin{equation}
\mathrm{SAM}= \frac{1}{N} \sum_{i=1}^N \arccos \left(\frac{\langle\hat{R}^{i}, R^{i}_{gt}\rangle}{\|\hat{R}^{i}\|_2\|R^{i}_{gt}\|_2}\right),
\end{equation}
where $N$ is the number of pixels.

\section{Spectral Illumination Transformer}
\begin{figure*}
  \centering
\includegraphics[width=.99\linewidth]{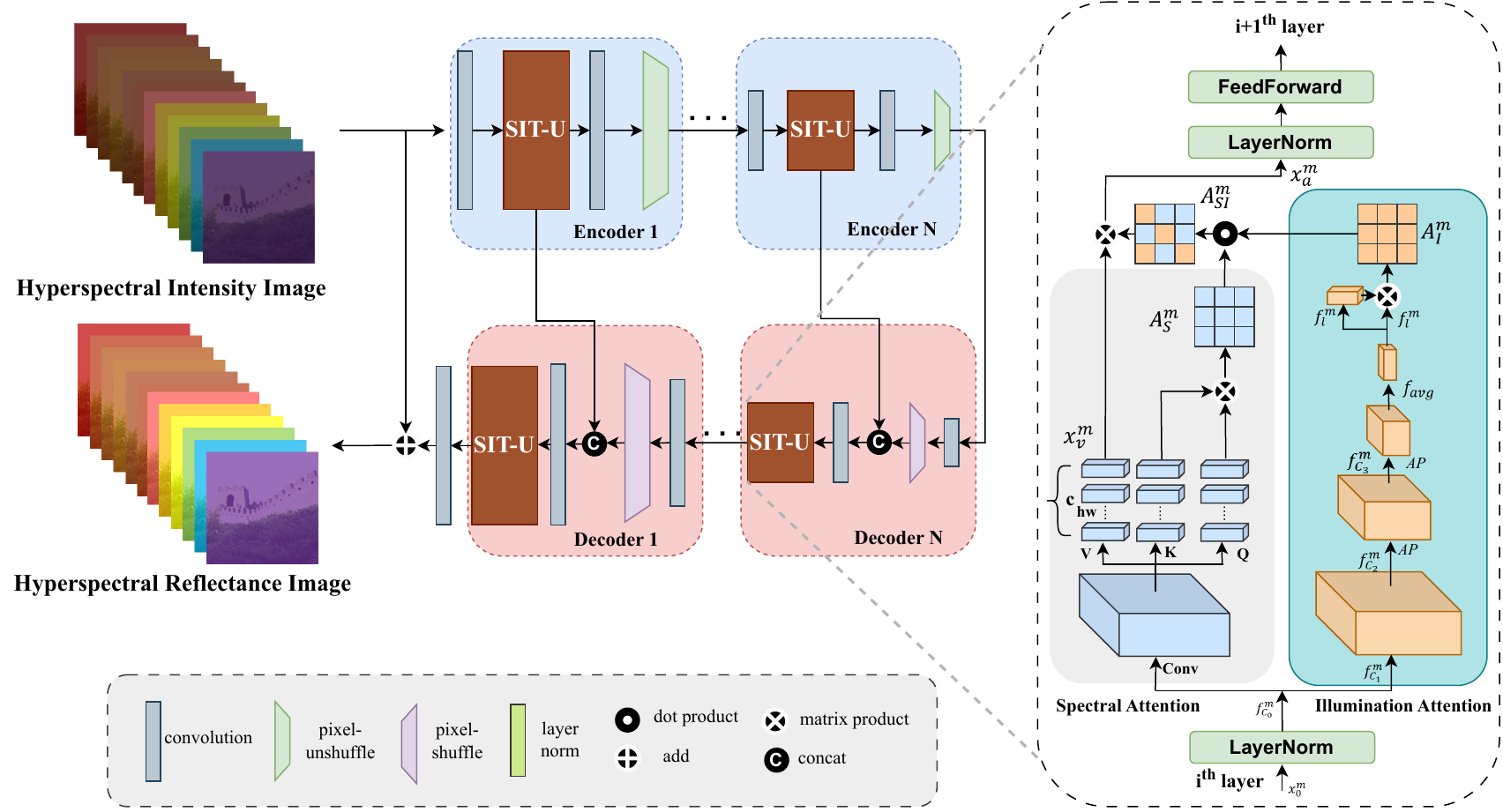}
  \caption{Structure of Spectral Illumination Transformer (SIT) framework and its unit (SIT-U) with Spectral and Illumination Attention branches for improved HSI calibration. The Illumination Attention branch mimics the Gray-World method to capture illumination features.}
  \label{fig-method-whole}
\end{figure*}



With the dataset ready, we propose the very first learning-based method which focuses on automatic illumination calibration for HSI: the \textit{Spectral illumination transformer (SIT)}. We consider SIT a baseline and will inspire future research.

\subsection{Framework Overview}

As shown in Fig.\ref{fig-method-whole}, the overall framework of our SIT is a U-shaped transformer proposed by {\cite{hcanet}} with residual connections. The framework is inspired by the commonly used U-Net for RGB images {{\cite{unet}}}, and later extended to the transformer by Wang~\etal \cite{uformer}. 
Hu~\etal\cite{hcanet} further extends it to HSI. The U-shaped encoder-decoder structure enables exploring more contextual information by increasing reception fields. And they use the spectral transformer to explore the relationship between channels.

However, as will be shown later by the experiments, this structure was not specially designed for HSI calibration and cannot properly capture the global illumination.
Therefore, we develop a novel spectral illumination transformer unit (SIT-U).

\subsection{Spectral Illumination Transformer Unit}

The SIT-U is inspired by the classic RGB white balancing algorithm Gray-World {\cite{grayworld}}. 
The Gray-World method assumes the global mean values of all pixels in each channel are the same (\eg 0.5).

As shown in Fig. \ref{fig-method-whole}, 
The SIT-U is an $M$-layer transformer. On the $m^{th}$ layer of SIT-U, the input feature $x^{m}_{0}$ of the $m^{th}$ layer of SIT-U  goes through the layer normalization $LN$ and a convolutional layer $f_{c_0}^{m}$ by
\begin{equation}
    x_{1}^{m} = f_{c_0}^{m}(LN(x_{0}^{m})), m = 1\cdots M.
\end{equation}
Then $x_{1}^{m}$ is injected to the two branches: spectral attention (SA) calculation and illumination attention (IA) calculation. For SA calculation, we maintain the same structure as HCANet\cite{hcanet}. We use SA to calculate the attention map $A_{S}^{m}$ and the attention ``value'' $x_{v}^{m}$ in the $m^{th}$ layer of SIT.
Instead of immediately multiplying $A_{S}^{m}$ with $x_{v}^{m}$ directly as HCANet, we design the illumination attention module to import the factors of illumination into the attention process, which will be introduced in detail in the following paragraph.



\paragraph{Illumination Attention}

Inspired by the Gray-World algorithm, we propose the Illumination Attention (IA).
Specifically, in the second branch ``IA'',  we use the illumination extractor to gradually capture global illumination features $x_{I}^{m}$.

First, it goes through the CNN layer $f_{c_{1}}^{m}$ keeping the spatial dimension by
\begin{equation}
   x^{m}_{2} = f_{c_1}^{m}(x^{m}_{1}).
\end{equation}
Then, we gradually extract local illumination features through down-scaling convolution and average pooling $AP$, by
 \begin{equation}
   x^{m}_{3} = AP (f_{c_2}^{m}(x^{m}_{2})),
   \end{equation}
    \begin{equation}
   x^{m}_{4} = AP (f_{c_3}^{m}(x^{m}_{3})),
\end{equation}
where the convolutional layer $f_{c,I_2}^{m}$ and $f_{c,I_3}^{m}$ have a stride of $t$ and the pooling size of $AP$ is $p$.
Finally, we calculate the global mean value in each channel by
 \begin{equation}
   x^{m}_{I} = f_{avg} (x^{m}_{4}),
\end{equation}
where $f_{avg}$ is used to scale down the feature map $x^{m}_{4}$ into a tensor $x^{m}_{I}$.
This averaging operation can explore the illumination of the entire image as the Gray-world\cite{grayworld} method.

Then, we linearly project the illumination feature into attentive hidden space using $f_{l}^{m}$ by 
\begin{equation}
    q_{I}^{m} = k_{I}^{m} = f_{l}^{m}(x_{I}^{m}),
\end{equation}
where $q_{I}^{m}$ and $k_{I}^{m}$ are the query and key of attention model.
The attention matrix $A_{I}^{m}$ in the IA branch of $m^{th}$ layer can be calculated by
\begin{equation}
    A_{I}^{m} = (q_{I}^{m})^{T} k_{I}^{m}.
\end{equation}
Through the IA branch, we explore the features of global illumination and calculate the other attention matrix $A_{I}^{m}$.

\paragraph{Spectral Illumination Attention}

Then we combine the two attention matrices $A_{S}^{m}$ and $A_{I}^{m}$ by 
\begin{equation}
    A_{SI}^{m} = \mathrm{softmax}(A_{S}^{m} \cdot A_{I}^{m}), 
\end{equation}
where the $A_{SI}^{m}$ is the final attention matrix in the spectral illumination attention.
Then, we calculate the output of spectral illumination attention $x_{a}^{m}$ by
\begin{equation}
    x_{a}^{m} = A_{SI}^{m} x_{v}^{m},
\end{equation}
where the $x_{v}^{m}$ is calculated in the spectral attention branch.

Then, as shown in Fig.\ref{fig-method-whole}, the output $x_{v}^{m}$ will go through layer normalization and feed-forward layers to calculate the final output of the $m^{th}$ layer of SIT-U.

\section{Experiments}
\label{sec:exp}

\begin{figure*}
  \centering
\includegraphics[width=\linewidth]{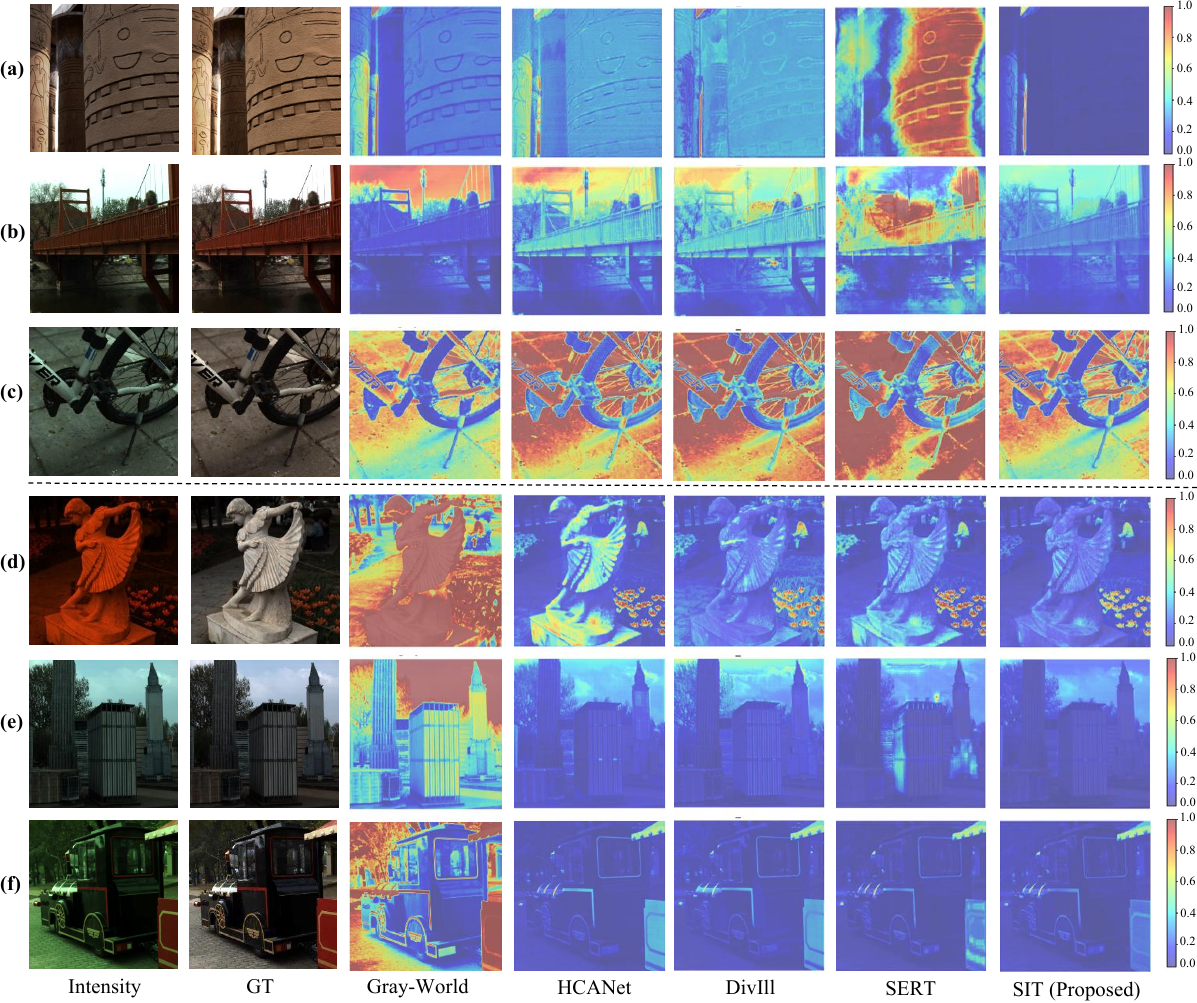}
  \caption{
  Visual Comparison of Absolute Error Using Heat Maps: The first two columns show the RGB rendering of the input and ground truth (GT) for visualization. Rows 1-3 display samples from BJTU-UVA, and Rows 4-6 show samples from BJTU-UVA-E.
  }
  \label{fig:vis}
\end{figure*}

\subsection{Experiment Settings}
The BJTU-UVA dataset is split into three sets: 535 for training, 114 for validation, and 116 for testing. Similarly, for the expansion BJTU-UVA-E: 5350 for training, 1140 for validation, and 1160 for testing.

Then we propose two tracks of benchmark. The first is the full spectrum test (400-1000nm, 204 channels); and the second is the 31-channel test, which is to align with existing HSI-enhancement dataset and methods {\cite{cave}\cite{havard}\cite{ntire2022}\cite{du2023}}. 

Because there are no existing learning-based methods to compare. We choose one classic baseline  Gray-world {\cite{grayworld}}; one recent indoor spectral restoration method DivIll \cite{indoorhsi} and two recent HSI denoising methods: SERT\cite{sert} and HCANet\cite{hcanet}.

All models in our experiments are trained on Nvidia 3090 GPUs. We randomly crop the image with a size of 256 and train with a batch size of 4.  We train our model and other models for 500 epochs on the  BJTU-UVA dataset and 50 epochs on the BJTU-UVA-E dataset.
We use L1 loss for optimizing pixel-level differences and Adam optimizer with an initial learning rate of $10^{-4}$.

\subsection{Benchmarks}
\paragraph{Comparisom on Full-Spectrum HSI}

\begin{table}
\centering
\renewcommand{\arraystretch}{1.2} 
\scriptsize
\setlength{\tabcolsep}{3 pt} 
\begin{tabular}{c|c|cccc|cccc}
\toprule
& & \multicolumn{4}{c|}{Val} & \multicolumn{4}{c}{Test} \\
\cmidrule(lr){3-6} \cmidrule(lr){7-10}
& & P$\uparrow$ & S $\downarrow$ & R$\downarrow$ & E$\downarrow$ & P$\uparrow$ & S $\downarrow$ & R$\downarrow$ & E$\downarrow$ \\
\midrule
\multirow{5}{*}{BJTU-UVA}
& GrayWorld\cite{grayworld} & 22.6  & 2.9 & 9.1 & 11.3 & 23.3 & 3.1 & 8.4 & 11.6 \\
& DivIll\cite{indoorhsi} & 23.5 & 5.5 & 8.3 & 10.2 & 24.1 & 6.1 & 7.6 & 10.0\\
& SERT\cite{sert} & 22.3 & 10.4 & 8.3 & 13.2& 22.8 & 11.1 & 7.9 & 13.5 \\
& HCANet\cite{hcanet} &  25.1& 4.4 & 6.8 & 9.1 & 25.7 & 3.6 & 6.4 & 9.7 \\
& \textbf{SIT}(proposed) & \textbf{26.1} & \textbf{2.9} & \textbf{6.2} & \textbf{8.7} & \textbf{26.3} & \textbf{3.1} & \textbf{5.8} & \textbf{9.6}\\
\midrule
\multirow{5}{*}{BJTU-UVA-E}
& GrayWorld\cite{grayworld} & 22.5 & 2.6 &  9.3& 11.5 & 23.2 & 2.8  & 8.6 & 11.7\\
& DivIll\cite{indoorhsi} & 35.2 & 3.1 & 2.2 & 2.7 & 35.2 & 3.4 & 2.2 & 3.0\\
& SERT\cite{sert} & 36.4 & 2.7 & 1.8 & 2.6 & 36.4 & 2.9 & 1.7 & 2.7 \\
& HCANet\cite{hcanet} & 37.9 & 2.1 & 1.4 & 2.0 & 37.3 & 2.5 & 1.5 & 2.3\\
& \textbf{SIT}(proposed) & \textbf{39.3} & \textbf{1.8} & \textbf{1.2} & \textbf{1.7} & \textbf{39.1} & \textbf{2.0} & \textbf{1.3} & \textbf{1.9}\\

\bottomrule
\end{tabular}
\caption{Automatic spectral calibration evaluation on the full-spectrum HSI:
PSNR(P), SAM(S), RMSE (R\%), and ERGAS (E\%).
}
\label{tab:exp-full}
\end{table}

Table \ref{tab:exp-full} presents the results. 
The first part shows the results on the original dataset.
We observe that the GrayWorld method shows relatively good performance, particularly with stable SAM scores.
Among deep learning-based approaches, DivIll, SERT, and HCANet improve on metrics like PSNR, RMSE, and ERGAS; however, they struggle to significantly enhance SAM scores.
In contrast, our proposed SIT method achieves the highest SAM scores across both tracks, with substantial improvements in other evaluation metrics as well. 

The lower part shows the results on the expanded dataset.
The GrayWorld method is not comparable to the other deep learning methods on all three tracks. 
Although deep learning methods improve their performance on all three tracks, our proposed method SIT still gets the best performance on all metrics.  

We visualize the experimental results in Fig. \ref{fig:vis}.
As can be seen, the proposed SIT is dedicated to this novel task and consistently outperforms existing methods. More results are provided in the supplementary materials.

\paragraph{Comparison on 31 Channel HSI}
\begin{table}
\centering
\renewcommand{\arraystretch}{1.2} 
\scriptsize
\setlength{\tabcolsep}{3 pt} 
\begin{tabular}{c|c|cccc|cccc}
\toprule
& & \multicolumn{4}{c|}{Val} & \multicolumn{4}{c}{Test} \\
\cmidrule(lr){3-6} \cmidrule(lr){7-10}
& & P$\uparrow$ & S $\downarrow$ & R$\downarrow$ & E$\downarrow$ & P$\uparrow$ & S $\downarrow$ & R$\downarrow$ & E$\downarrow$ \\

\midrule
\multirow{5}{*}{BJTU-UVA} 
& GrayWorld\cite{grayworld} & 25.2 & 3.3 & 5.8 & 9.6 & 24.2 & 4.1 & 6.5 & 5.2\\
& DivIll\cite{indoorhsi} & 22.4 & 12.7  & 7.7 & 14.3 & 21.7 & 13.3 & 7.5 & 16.5\\
& SERT\cite{sert} & 23.6 & 10.1 & 6.6  & 13.4 & 23.0 & 10.3 & 6.4 & 15.2 \\
& HCANet\cite{hcanet} & 26.4 & 3.5 & 5.6 & 9.1 & 26.3 & 3.4 & 5.0 & 11.0\\
& \textbf{SIT}(proposed) & \textbf{26.8}  & \textbf{2.9} & \textbf{5.3} & \textbf{9.2} & \textbf{26.7} & \textbf{2.7} & \textbf{5.1} & \textbf{11.7} \\
\midrule
\multirow{5}{*}{BJTU-UVA-E} 
& GrayWorld\cite{grayworld} & 25.3 & 3.1 & 5.8 & 9.5 & 24.2 & 3.6 &  6.5 & 14.1\\
& DivIll\cite{indoorhsi} & 37.2 & 2.6 & 1.6 & 3.0 & 37.8 & 2.7 &1.3  & 2.9\\
& SERT\cite{sert} & 36.6 & 3.1 & 1.7 & 3.3 & 37.0 & 3.3 & 1.5 & 3.6 \\
& HCANet\cite{hcanet} & 37.8 & 2.4 & 1.3 & 2.5 & 38.1 & 2.5 & 1.2 & 2.8\\
& \textbf{SIT}(proposed) & \textbf{40.3} & \textbf{1.8} & \textbf{1.0} & \textbf{1.8} & \textbf{41.1} & \textbf{1.9} & \textbf{0.8} & \textbf{1.9}\\

\bottomrule
\end{tabular}
\caption{Automatic spectral calibration evaluation on 31 Channel HSI:PSNR(P), SAM(S), RMSE (R\%), and ERGAS (E\%).}
\label{tab:exp-standard}
\end{table}

Table \ref{tab:exp-standard} shows the performance on 31-channel HSI data. The results align with those from the full-spectrum experiments, demonstrating that the proposed SIT method achieves the best performance across all settings. Notably, all methods perform better than in the full-spectrum experiments, likely due to the reduced number of channels simplifying the tasks.

\paragraph{Discussions}

It is worth noticing for the original datasets, which are all-natural illuminations, GrayWorld is more reliable than some of the learning-based methods; it is reasonable because the "gray-world" assumption is an assumption on the illumination of natural images. As for the expansion dataset, when the illuminations are not directly captured from the natural scenes, the performance drops significantly.

It is also noticed that learning-based methods perform significantly better on the expansion dataset than on the original dataset, whereas Grayworld does not have such improvements. This is probably because: first the expansion dataset is much larger; second the diversified illuminations allow the learning-based methods to better learn and model the illuminations.


\subsection{Ablation Study} 

\begin{table}

\centering
\renewcommand{\arraystretch}{1.2} 
\scriptsize
 \setlength{\tabcolsep}{5pt} 
\begin{tabular}{c c|cccc|cccc}
\toprule
& & \multicolumn{4}{c|}{Validation} & \multicolumn{4}{c}{Test} \\
\cmidrule(lr){3-6} \cmidrule(lr){7-10}
& & P $\uparrow$ & S $\downarrow$ & R $\downarrow$ & E $\downarrow$ & P $\uparrow$ & S $\downarrow$ & R $\downarrow$ & E $\downarrow$ \\
\midrule
& W/O SA \& IA        &  21.4  & 11.5  & 9.3   & 15.1   & 22.1   & 8.5  & 11.9   & 15.6 \\
& IA         & 25.5   & 4.1   & 6.3   & 9.2   & 25.8   & 4.3   & 5.9   & 9.6   \\
& SA         & 25.1  & 4.4   & 6.8   & 9.1   & 25.7  & 3.6   & 6.4   & 9.7   \\
& IA \& SA & \textbf{26.1}  & \textbf{2.9}   & \textbf{6.2}   & \textbf{8.7}   & \textbf{26.3}  & \textbf{3.1}   &\textbf{5.8}  & \textbf{9.6}   \\
\bottomrule
\end{tabular}
\caption{Ablation Study on BJTU-UVA Full-Spectrum Data: Performance Comparison of Different Attention Configurations (Spectral Attention (SA) and Illumination Attention (IA))}
\label{tab:ablation}
\end{table}



We study how useful the proposed illumination attention (IA) module is.
Four experimental setups are evaluated: ``W/O SA $\&$ IA'', where only CNN layers are used; Only ``SA'', is a setting which falls back to HCANet {\cite{hcanet}}; ``IA'', where attention similarity calculation uses only IA; and ``IA $\&$ SA'', our complete SIT approach, where the attention maps of IA and SA are combined in each layer. 

As shown in Table \ref{tab:ablation}, IA can significantly improve the performance with and without the presence of SA. When both SA and IA are present, the performance is optimal. Besides, it is interesting to find that although the structure is different, the performance of SA and IA are similar when working separately.

\begin{figure}
  \centering
\includegraphics[width=.99\linewidth]{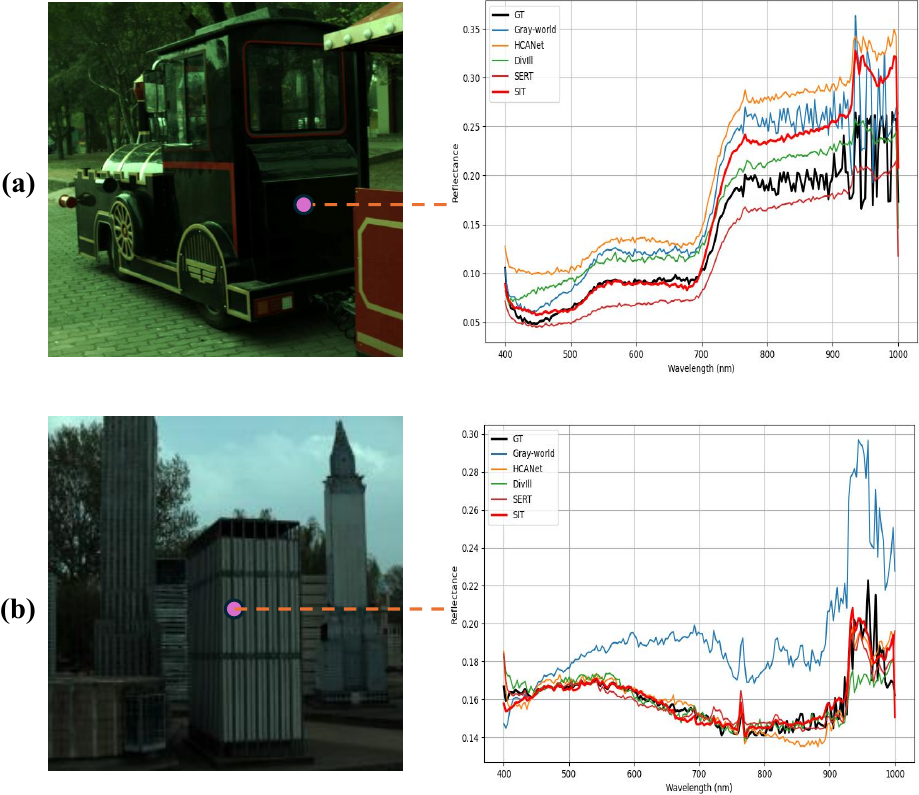}
  \caption{Challenging Cases: Comparison of Calibrated Spectra by Different Methods}
  \label{fig-chal}
\end{figure}

\subsection{Challenging Cases}

As can be seen in Fig.~\ref{fig-chal}, although the proposed SIT performs relatively better than other methods, it still faces challenges in low-light scenes and areas. 
This challenge is more significant in the infrared bands, where the illumination is generally less intense than the visible bands. 
We consider handling low-light scenes as a potential future work.

\section{Conclusion}


This paper introduces a learning-based approach for the automatic calibration of hyperspectral images (HSIs). A large-scale HSI calibration dataset, BJTU-UVA, has been developed, comprising 765 high-quality HSI pairs that encompass diverse natural scenes and illumination conditions. This dataset is further augmented to 7,650 pairs by incorporating 10 distinct, physically measured illuminations. We propose a spectral illumination transformer (SIT) with an illumination attention module, achieving state-of-the-art performance, as demonstrated by extensive benchmarks. An ablation study highlights the effectiveness of the illumination attention module. The benchmark also reveals limitations in low-light areas across both visible and infrared bands, suggesting a direction for future work.

\section*{Acknowledgement}
This work was completed during the author's visit to the University of Amsterdam (UVA), supported by the China Scholarship Council (CSC). The authors would like to thank the CSC for providing funding and support, and the University of Amsterdam for offering a collaborative and inspiring research environment.

\section*{Dataset Usage License}

The \textbf{BJTU-UVA} dataset, jointly created by Beijing Jiaotong University (BJTU) and the University of Amsterdam (UVA), is made available for academic and research purposes under the following terms:

\begin{enumerate}
    \item \textbf{Permitted Uses:}
    \begin{itemize}
        \item The dataset may be used solely for non-commercial research, development, and educational purposes.
        \item Users are permitted to analyze, process, and modify the dataset to suit their research needs, provided any modifications are not redistributed without prior permission.
        \item Proper citation of the dataset and related publications is required in any academic or research work utilizing the dataset. Below is the BibTeX entry:
\noindent\parbox{\textwidth}{%
\small 
\texttt{@misc\{du2024spectral,\\
\ \ \ \ title=\{Automatic Spectral Calibration \\
of Hyperspectral Images: \\
Method, Dataset and Benchmark\},\\
\ \ \ \ author=\{Zhuoran Du and Shaodi You \\
and Cheng Cheng and Shikui Wei\},\\
\ \ \ \ year=\{2024\},\\
\ \ \ \ eprint=\{2412.14925\},\\
\ \ \ \ archivePrefix=\{arXiv\},\\
\ \ \ \ primaryClass=\{cs.CV\},\\
\ \ \ \ url=\{https://arxiv.org/abs/2412.14925\},\\
\}
}
}
    \end{itemize}

    \item \textbf{Prohibited Uses:}
    \begin{itemize}
        \item The dataset may not be used for commercial purposes, including integration into products for sale or other commercial activities.
        \item Redistribution, sublicensing, or sharing of the dataset or derivative works without explicit written permission from the authors is strictly prohibited.
        \item Any usage violating ethical, legal, or regulatory standards is prohibited.
    \end{itemize}


    \item \textbf{No Warranty:}  
    The dataset is provided "as is" without any warranties, express or implied, including but not limited to accuracy, completeness, or fitness for any particular purpose. The authors are not liable for any damage or loss arising from the use of the dataset.

    \item \textbf{Ethical Considerations:}  
    Users must adhere to ethical research practices, particularly in terms of data privacy, responsible research, and compliance with relevant laws and regulations.

    \item \textbf{Access and Updates:}  
    Access to the dataset is provided under these terms. The authors reserve the right to modify these terms or revoke access at any time.
\end{enumerate}

By using the \textbf{BJTU-UVA} dataset, you agree to comply with the terms outlined above.

\clearpage
{
    \small
    \bibliographystyle{ieeenat_fullname}
    \bibliography{main}
}

\clearpage
\maketitlesupplementary

We present detailed benchmarks, including separate evaluations for visible light and near-infrared (NIR) spectra. Additionally, we provide illumination-specific benchmarks for the expansion dataset. To enhance understanding, we include more visual results from both the original and expansion datasets. Lastly, we offer a comprehensive comparison of existing hyperspectral imaging (HSI) datasets for natural scenes.

\section{Detailed Evaluation of Calibration Methods: Visible and Near-Infrared Ranges}
\label{sec:vis-nir}
To further verify the quality of the spectral calibration, we evaluate the calibrated HSIs from the visible (400-700nm) and near-infrared (700-1000nm) perspectives. We conduct experiments on both the BJTU-UVA and BJTU-UVA-E datasets.
\paragraph{Comparison on the BJTU-UVA}


Table \ref{tab:vis_nir_comparison} presents the results. Compared to other learning-based methods, our approach demonstrates a significant advantage across all evaluation metrics in both the visible and near-infrared (NIR) ranges.

In the visible range, our proposed SIT model achieves superior scores in PSNR, RMSE, and ERGAS. Notably, the Gray-World method excels in visible range calibration, achieving the best SAM value. Leveraging the strengths of the Gray-World approach, our SIT model is designed to more effectively capture illumination compared to other learning-based methods.

In the NIR range, the performance of all methods declines considerably, underscoring the inherent challenges of this spectrum compared to the visible range. Despite this, our method consistently achieves the best scores across all metrics, reaffirming its robustness. The inclusion of the NIR spectrum in our dataset emphasizes the need to address these challenges and drive further advancements in this area.

\paragraph{Comparison on the BJTU-UVA-E}

On the BJTU-UVA-E dataset, our proposed SIT model consistently achieved the best scores across all evaluation metrics. In this context, the Gray-World method was not comparable to the learning-based approaches. Similarly, a significant performance drop was observed in the NIR range, highlighting the persistent challenges associated with this spectrum. Addressing the calibration limitations of the NIR range remains an open challenge for future research.

\begin{table*}
\centering
\renewcommand{\arraystretch}{1.2} 
\scriptsize
\setlength{\tabcolsep}{5 pt} 
\begin{tabular}{c|c|cccc|cccc|cccc|cccc}
\toprule
& & \multicolumn{8}{c|}{Val} & \multicolumn{8}{c}{Test} \\
\cmidrule(lr){3-10} \cmidrule(lr){11-18}
& & \multicolumn{4}{c|}{VIS} & \multicolumn{4}{c|}{NIR} & \multicolumn{4}{c|}{VIS} & \multicolumn{4}{c}{NIR} \\
\cmidrule(lr){3-6} \cmidrule(lr){7-10} \cmidrule(lr){11-14} \cmidrule(lr){15-18}
& & P$\uparrow$ & S $\downarrow$ & R$\downarrow$ & E$\downarrow$ & P$\uparrow$ & S $\downarrow$ & R$\downarrow$ & E$\downarrow$ 
& P$\uparrow$ & S $\downarrow$ & R$\downarrow$ & E$\downarrow$ & P$\uparrow$ & S $\downarrow$ & R$\downarrow$ & E$\downarrow$ \\
\midrule
\multirow{5}{*}{BJTU-UVA}
& GrayWorld\cite{grayworld} & 26.0 & \textbf{2.5} & 5.4 & 8.9 & 21.4 & 3.1 & 10.8 & 10.5 & 25.1 & \textbf{2.8} & 6.0 & 13.3 & 22.3 & 3.3 & 9.4&9.8 \\
& DivIll\cite{indoorhsi} & 25.6 & 5.7 & 6.3 & 9.6 & 21.7 & 5.3 & 9.6 & 9.8 & 25.0 & 6.2 & 6.0 & 10.6 & 22.5 & 5.9 & 8.7 & 9.2 \\
& SERT\cite{sert} & 24.1 & 10.5 & 6.3 & 12.4 & 20.2 & 11.2 & 9.9 & 13.2 & 23.6 & 10.6 & 6.1 & 14.2 & 20.8 & 11.5 & 9.1 & 12.7 \\
& HCANet\cite{hcanet} & 27.5 & 4.5 & 5.0 & 8.2 & 23.6 & 4.1 & 7.7 & 8.9 & 26.6 & 4.5 & 5.0 & 10.3 & 23.9 & 4.3 & 7.3 & 9.1 \\
& \textbf{SIT}(proposed) & \textbf{28.5} & 2.9 & \textbf{4.5} & \textbf{8.1} & \textbf{24.1} & \textbf{3.0} & \textbf{7.3} & \textbf{8.7} & \textbf{27.4} & 3.1 & \textbf{4.6} & \textbf{10.0} & \textbf{24.5} & \textbf{3.3} & \textbf{6.6} & \textbf{8.9} \\
\midrule
\multirow{5}{*}{BJTU-UVA-E}
& GrayWorld\cite{grayworld} & 25.9 & 2.2 & 5.7 & 9.1 & 21.3 & 3.0 & 11.1 & 10.6 & 25.0 & 2.2 & 6.2 & 13.6 & 22.4 & 3.2 & 9.5 & 9.8 \\
& DivIll\cite{indoorhsi}& 36.6 & 3.3 & 1.7 & 2.9 & 33.4 & 3.3 & 2.5 & 2.6 & 37.0 & 3.7 & 1.5 & 3.1 & 33.1 & 3.7 & 2.6 & 2.9\\
& SERT\cite{sert} & 37.3 & 3.1 & 1.8 & 2.9 & 33.3 & 3.1 & 2.6 & 3.0 & 37.8 & 3.3 & 1.5 & 3.0 & 33.1 & 3.3 & 2.5 & 3.0 \\
& HCANet\cite{hcanet}  & 40.6 & 2.2 & 1.0 & 1.9 & 35.3 & 2.2 & 1.8 & 2.1 & 41.1 & 2.4 & 0.9 & 2.1 & 34.7 & 2.4 & 1.9 & 2.3 \\
& \textbf{SIT}(proposed) & \textbf{42.5} & \textbf{1.7} & \textbf{0.8} & \textbf{1.5} & \textbf{37.2} & \textbf{1.7} & \textbf{1.5} & \textbf{1.7} & \textbf{43.2} & \textbf{1.8} & \textbf{0.7} & \textbf{1.6} & \textbf{36.5} & \textbf{1.8} & \textbf{1.6} & \textbf{1.9} \\

\bottomrule
\end{tabular}
\caption{Automatic spectral calibration evaluation on both visible-spectrum (VIS, 400nm-700nm) and near-infrared (NIR, 700nm-1000nm) HSI: PSNR(P), SAM(S), RMSE (R\%), and ERGAS (E\%).}
\label{tab:vis_nir_comparison}
\end{table*}

\section{Evaluation on Different Illumination Conditions of the BJTU-UVA-E Dataset}




To evaluate model performance under varying illumination conditions, we analyzed the PSNR calibration results across ten different illumination settings. The definitions and spectra of these illumination settings are illustrated in Fig.~\ref{illu-record}, and the results are summarized in Table \ref{tab:combined-results}.

The Gray-World method, which relies on linear functions of averaging and division, produces uniform predictions across different illumination settings, making it incomparable to other methods in this context. By contrast, our proposed SIT method consistently outperforms all competitors across all illumination settings.

Focusing on the five natural illuminations, we observe that dusk (du) and shadowy (sh) conditions pose greater calibration challenges due to their low-light environments. Even in these difficult settings, our method achieves the highest PSNR values, with 37.1 in "du" and 36.3 in "sh" for the NIR range.

For the five color-filtered illuminations, the red-filtered illumination presents the most significant challenge, as it limits visible light intensity between 400nm and 500nm. Despite this, our method maintains its superior performance.


\begin{table*}[tbp]
\centering
\scriptsize 
\setlength{\tabcolsep}{3pt} 
\renewcommand{\arraystretch}{1.1} 
\begin{tabular}{l|c*{10}{c}|c*{10}{c}}
\toprule
 &  & \multicolumn{10}{c|}{\textbf{Validation Set}} & \multicolumn{10}{c}{\textbf{Test Set}} \\
 
 &  & \textbf{sd} & \textbf{cd} & \textbf{rd} & \textbf{du} & \textbf{sh} & \textbf{bl} & \textbf{re} & \textbf{ye} & \textbf{gr} & \textbf{pu} & \textbf{sd} & \textbf{cd} & \textbf{rd} & \textbf{du} & \textbf{sh} & \textbf{bl} & \textbf{re} & \textbf{ye} & \textbf{gr} & \textbf{pu} \\
\midrule
\multirow{3}{*}{\textbf{GrayWorld\cite{grayworld}}} & \textbf{vis} & 25.9 & 25.9 & 25.9 & 25.9 & 25.9 & 25.9 & 25.9 & 25.9 & 25.9 & 25.9 & 25.0 & 25.0& 25.0& 25.0& 25.0& 25.0& 25.0& 25.0& 25.0& 25.0  \\
 & \textbf{nir} & 21.3 & 21.3 & 21.3 & 21.3 & 21.3 & 21.3 & 21.3 & 21.3 & 21.3 & 21.3 & 22.4 & 22.4& 22.4& 22.4& 22.4& 22.4& 22.4& 22.4& 22.4& 22.4  \\
 & \textbf{whole} & 22.5 & 22.5 & 22.5 & 22.5 & 22.5 & 22.5 & 22.5 & 22.5 & 22.5 & 22.5 & 23.2 & 23.2& 23.2& 23.2& 23.2& 23.2& 23.2& 23.2& 23.2& 23.2  \\
\midrule
\multirow{3}{*}{\textbf{DivIll\cite{indoorhsi}}} & \textbf{vis} & 37.5 & 38.9 & 39.1 & 36.5 & 31.8 & 38.1 & 34.2 & 38.1 & 36.9 & 38.6 & 37.9 & 39.8 & 39.6 & 36.5 & 30.8 & 38.8 & 34.7 & 38.4 & 37.7 & 39.3  \\
 & \textbf{nir} & 34.0 & 32.1 & 33.9 & 32.5 & 29.4 & 34.8 & 35.1 & 33.8 & 35.8 & 35.8 & 33.4 & 32.9 & 33.5 & 32.0 & 28.6 & 34.5 & 34.8 & 33.6 & 35.3 & 35.4  \\
 & \textbf{whole} & 35.9 & 34.8 & 36.2 & 34.7 & 31.0 & 36.7 & 35.2 & 36.0 & 36.9 & 37.5 & 35.7 & 35.9 & 36.1 & 34.5 & 30.4 & 36.7 & 35.5 & 36.0 & 36.9 & 37.6 \\
\midrule
\multirow{3}{*}{\textbf{HCANet\cite{hcanet}}} & \textbf{vis} & 43.1 & 43.0 & 42.6 & 40.8 & 41.4 & 40.8 & 34.5 & 39.6 & 38.4 & 41.8 & 43.9 & 44.4 & 43.5 & 41.8 & 42.9 & 41.7 & 35.2 & 40.1 & 38.9 & 42.2  \\
 & \textbf{nir} & 35.8 & 34.4 & 34.6 & 34.3 & 33.8 & 36.0 & 36.2 & 35.7 & 36.2 & 36.2 & 35.3 & 34.6 & 34.0 & 34.2 & 33.9 & 35.6 & 35.8 & 35.3 & 35.8 & 35.8  \\
 & \textbf{whole} & 38.6 & 37.4 & 37.4 & 37.0 & 36.7 & 38.3 & 35.7 & 37.7 & 37.6 & 38.6 & 38.4 & 37.9 & 37.1 & 37.2 & 37.1 & 38.3 & 36.1 & 37.6 & 37.6 & 38.5 \\
\midrule
\multirow{3}{*}{\textbf{SERT\cite{sert}}} & \textbf{vis} & 40.5 & 42.6 & 38.2 & 36.1 & 28.1 & 38.0 & 35.2 & 38.8 & 38.1 & 40.7 & 41.1 & 43.4 & 38.7 & 36.4 & 29.6 & 38.4 & 35.2 & 39.4 & 38.4 & 41.2 \\
 & \textbf{nir} & 32.9 & 34.3 & 32.2 & 31.4 & 25.4 & 35.4 & 35.9 & 36.2 & 35.7 & 36.2 & 32.4 & 34.6 & 32.2 & 31.1 & 26.0 & 35.0 & 35.5 & 35.5 & 35.4 & 35.9 \\
 & \textbf{whole} & 35.8 & 37.3 & 34.8 & 33.8 & 26.9 & 37.1 & 36.1 & 37.7 & 37.2 & 38.5 & 35.7 & 37.8 & 35.1 & 33.9 & 28.1 & 37.1 & 36.1 & 37.5 & 37.3 & 38.4 \\
\midrule
\multirow{3}{*}{\textbf{SIT}} & \textbf{vis} & \textbf{44.5} & \textbf{47.7} & \textbf{46.2} & \textbf{42.7} & \textbf{43.9} & \textbf{42.5} & \textbf{36.4} & \textbf{40.3} & \textbf{39.3} & \textbf{44.8} & \textbf{45.3} & \textbf{48.4} & \textbf{47.2} & \textbf{43.1} & \textbf{44.8} & \textbf{42.8} & \textbf{36.8} & \textbf{41.0} & \textbf{40.0} & \textbf{45.5}  \\
 & \textbf{nir} & \textbf{37.4} & \textbf{37.6} & \textbf{37.4} & \textbf{37.1} & \textbf{36.3} & \textbf{37.8} & \textbf{38.0} & \textbf{37.7} & \textbf{37.9} & \textbf{38.3} & \textbf{36.5} & \textbf{37.1} & \textbf{36.8} & \textbf{36.4} & \textbf{35.7} & \textbf{37.0} & \textbf{37.2} & \textbf{36.9} & \textbf{37.0} & \textbf{37.6}  \\
 & \textbf{whole} & \textbf{40.2} & \textbf{40.8} & \textbf{40.3} & \textbf{39.6} & \textbf{39.1} & \textbf{40.0} & \textbf{37.6} & \textbf{39.2} & \textbf{38.9} & \textbf{41.0} & \textbf{39.7} & \textbf{40.5} & \textbf{40.1} & \textbf{39.2} & \textbf{38.9} & \textbf{39.6} & \textbf{37.5} & \textbf{39.1} & \textbf{38.8} & \textbf{40.6} \\

\bottomrule
\end{tabular}
\caption{Calibration results (PSNR) of visible, near-infrared, and full-range on the validation and test sets of BJTU-UVA-E.}
\label{tab:combined-results}
\end{table*}

\section{More Qualitative Analysis}

\begin{figure*}[p]
  \centering
\includegraphics[width=0.99\linewidth]{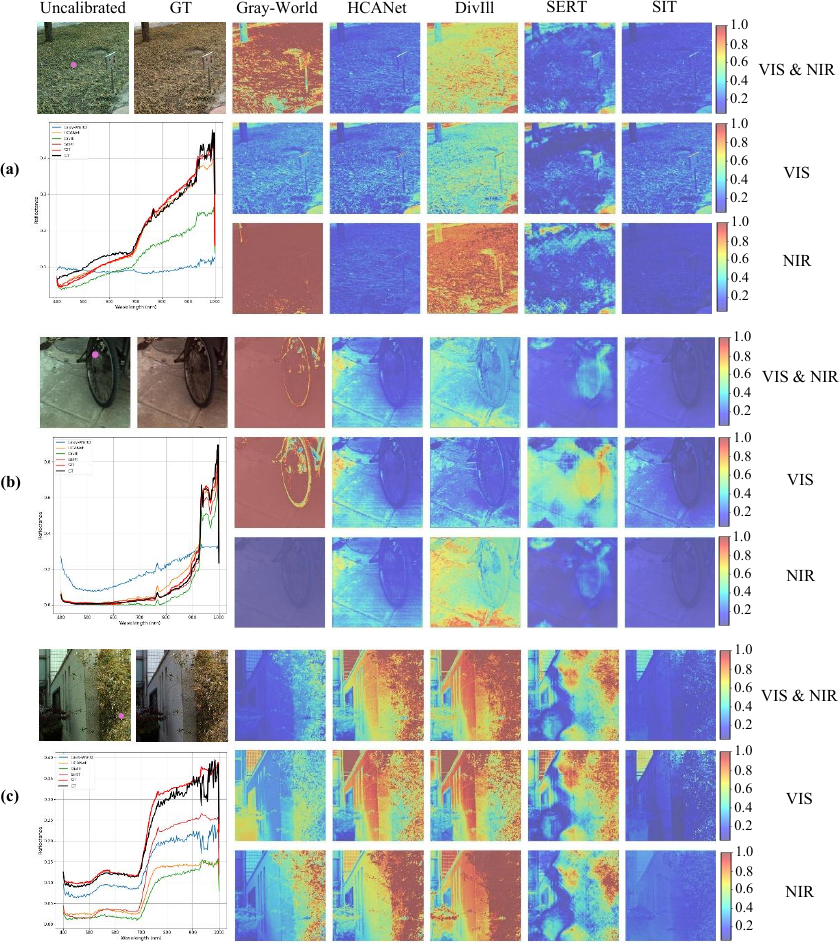}
  \caption{
  Visual Comparison of Absolute Error Using Heat Maps on BJTU-UVA. First row: Uncalibrated HSI, ground truth, and error maps (full spectrum) for five methods. Second and third rows: Error maps for VIS and NIR ranges. Bottom-left: Pixel spectra comparison across methods.
  }
  \label{fig:sup-real-1}
\end{figure*}
\begin{figure*}[tbp]
  \centering
\includegraphics[width=0.99\linewidth]{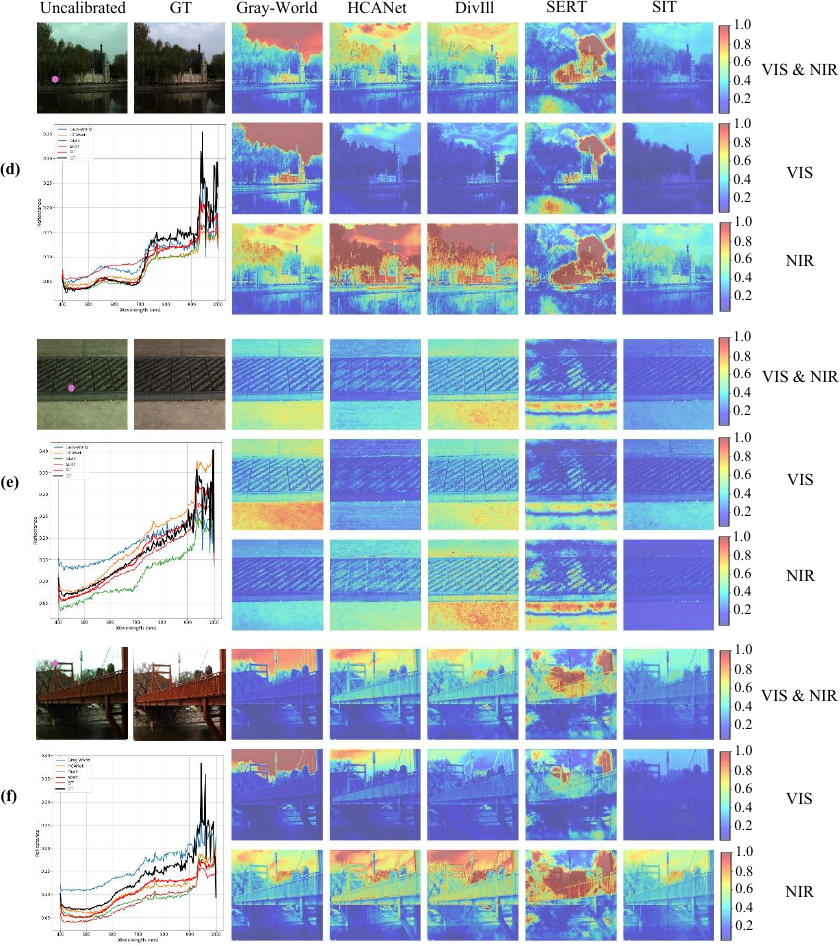}
  \caption{
  Visual Comparison of Absolute Error Using Heat Maps on BJTU-UVA.
  }
  \label{fig:sup-real-2}
\end{figure*}
\begin{figure*}[p]
  \centering
\includegraphics[width=0.99\linewidth]{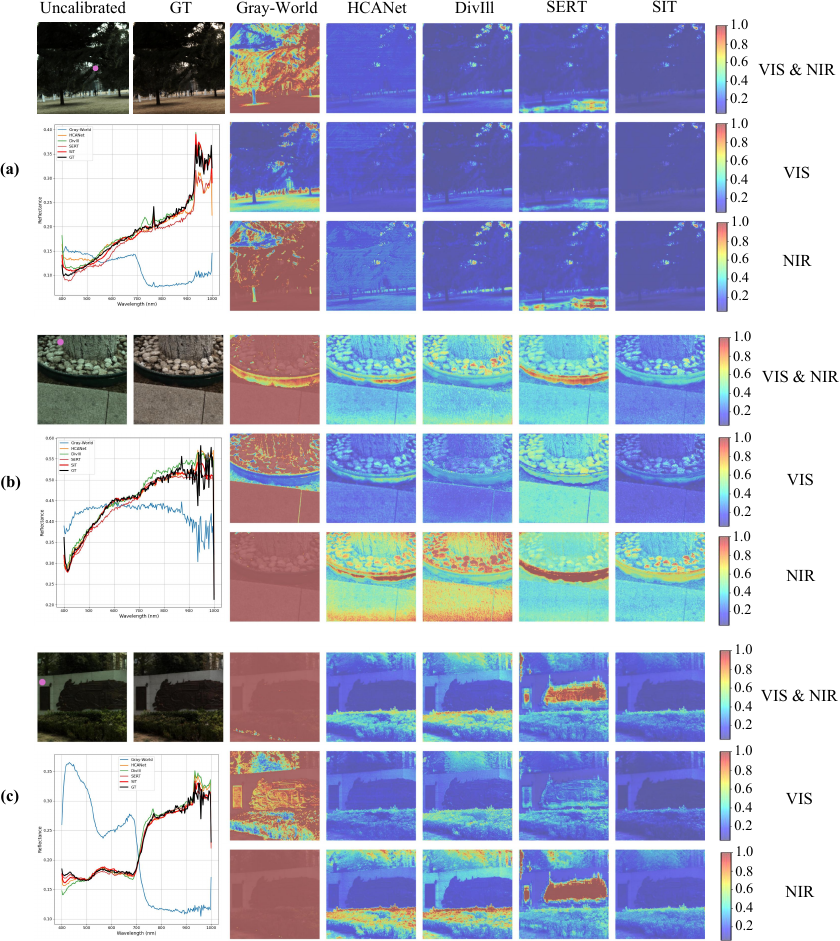}
  \caption{
  Visual Comparison of Absolute Error Using Heat Maps on BJTU-UVA-E.
  }
  \label{fig:sup-sim-1}
\end{figure*}
\begin{figure*}[p]
  \centering
\includegraphics[width=0.99\linewidth]{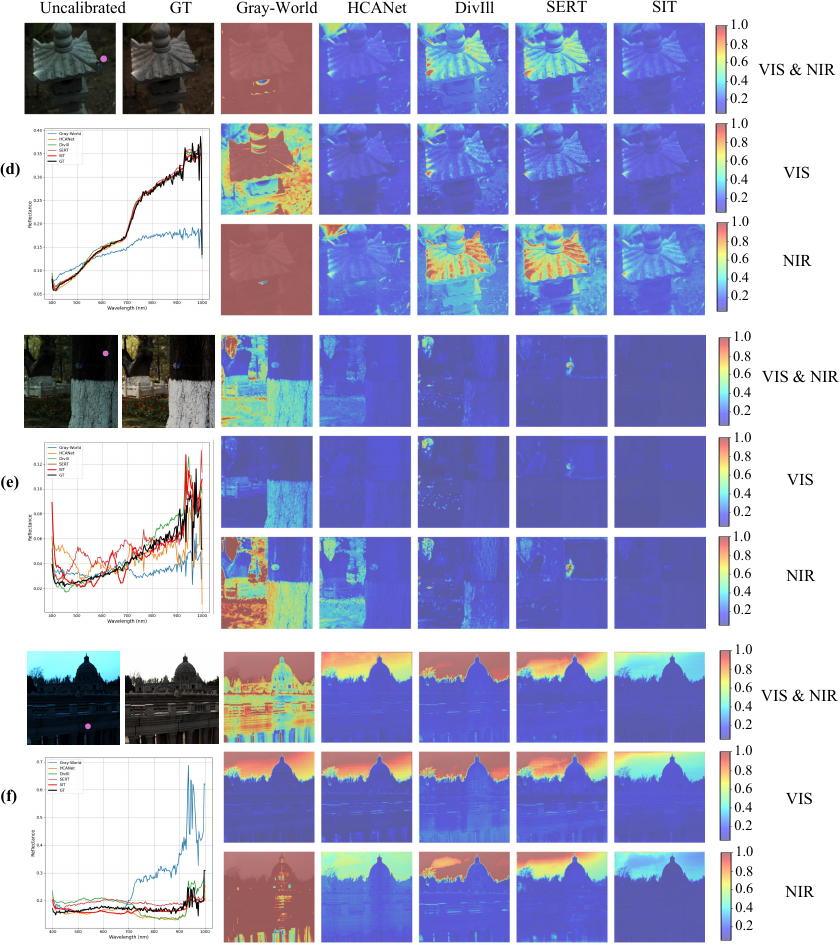}
  \caption{
  Visual Comparison of Absolute Error Using Heat Maps on BJTU-UVA-E.
  }
  \label{fig:sup-sim-2}
\end{figure*}

We present six examples from the BJTU-UVA dataset in Fig.~\ref{fig:sup-real-1} and Fig.~\ref{fig:sup-real-2}, and six additional examples in Fig.~\ref{fig:sup-sim-1} and Fig.~\ref{fig:sup-sim-2}. Each example includes the following evaluations:

\begin{itemize}
    \item \textbf{First row:} The first two images represent the uncalibrated HSI and the ground truth for the calibrated HSI, both rendered as RGB images for visualization. Columns 3 to 7 display error heat maps of the calibration results from five methods across the full spectral range.
    \item \textbf{Second and third rows:} These rows visualize the error heat maps specifically for the visible and near-infrared spectral ranges, respectively.
    \item \textbf{Bottom-left corner:} The spectrum of a selected pixel is plotted, showing the output from various calibration methods for comparison.
\end{itemize}

For the BJTU-UVA heat maps, we use a threshold of 0.14, derived from the global 95th percentile of a randomly selected subset. For the BJTU-UVA-E heat maps, we set the threshold to 0.03, based on the global 85th percentile of a similar subset, to enhance the clarity of the comparisons.




\begin{table*}[tbp]
\centering
\tabcolsep 0.05 in \scriptsize

\begin{tabular}{cccccccccc}
\hline
Dataset & Bands & Images & Spectral Range & Spatial Resolution & Denoising & Super-resolution & Spectral Recovery & Spectral Calibration & Year \\
\hline
CAVE [35]   & 31 & 32 & 400-700 & 512$\times$512 &\checkmark &\checkmark&\checkmark& & 2010 \\
Harvard [36]  & 31 & 50 & 400-700 & 1392$\times$1040 &\checkmark&\checkmark&\checkmark&& 2011 \\
NUS [37]  & 31 & 64 & 400-700 & 1392$\times$1300 &\checkmark&\checkmark&\checkmark& & 2014 \\
ICVL [38]  & 31 & 201 & 400-700 & 1392$\times$1300 &\checkmark&\checkmark&\checkmark&& 2016 \\
NTIRE'18 [39]  & 31 & 256 & 400-700 & 480$\times$512 & &&\checkmark&& 2018 \\
NTIRE'20 [40]  & 31 & 510 & 400-700 & 480$\times$512 &&&\checkmark& & 2020 \\
NTIRE'22 [41]  & 31 & 1000 & 400-700 & 480$\times$512 &&&\checkmark& &  2022 \\
FHRS [42] & 31 & 607 & 400-700 & 512$\times$512 &&&\checkmark& & 2023 \\
\hline
\textbf{BJTU-UVA} & \textbf{204} & \textbf{765 (7650)} & \textbf{400-1000} & \textbf{512$\times$512} &\checkmark*&\checkmark*&\checkmark*& \checkmark & \textbf{2024} \\
\hline
\end{tabular}
\caption{Comparison of public hyperspectral image datasets. 
Only the proposed dataset BJTU-UVA is suitable for auto-calibration because other datasets don't provide raw data. 
Although this paper focuses on auto-calibration, other tasks are also possible by the proposed dataset, which will be released later. 7650 is the size of the expansion dataset.
}
\label{tab:datasets}
\end{table*}
\paragraph{Qualitative Analysis on BJTU-UVA}

As illustrated in Fig.~\ref{fig:sup-real-1}(c), the calibration errors of our proposed SIT method are the smallest across all regions, particularly for pure blue areas. In contrast, other methods struggle to achieve accurate calibration, especially in the right regions containing more red areas.  
The spectral analysis further confirms that the calibrated spectrum curve produced by SIT exhibits the highest similarity to the ground truth.  
From the visualization results in the visible (VIS) and near-infrared (NIR) spectral ranges, it is evident that the error maps of NIR from other methods display significantly more red areas compared to VIS. In contrast, the heatmap generated by SIT remains predominantly blue, demonstrating its ability to effectively handle calibration in the near-infrared range.  
In the remaining five examples, our method consistently achieves the best results.

\paragraph{Qualitative Analysis on BJTU-UVA-E}

Similarly, our SIT method outperforms other approaches on the BJTU-UVA-E dataset. For instance, in Fig.~\ref{fig:sup-sim-1}(c), the Gray-World method performs poorly, with a calibrated spectrum that deviates significantly from the ground truth.  
The shrubbery regions with shadow in this example present a particular challenge, especially in the NIR range, yet the SIT method achieves near-perfect calibration in these areas.  
Across other examples, it is apparent that the Gray-World method consistently underperforms on the BJTU-UVA-E dataset, while SIT maintains superior results.





\section{Comparison with Existing Hyperspectral Datasets for Natural Scenes}

As shown in Table~\ref{tab:datasets}, the rapid development of hyperspectral imaging devices has shifted research focus from traditional remote sensing to natural environments on the ground.  
In natural scenes, several hyperspectral datasets have been proposed for restoration tasks, including denoising \cite{sert}, spatial super-resolution~[43], and spectral super-resolution~[42]. These datasets typically focus on the visible spectrum, ranging from 400~nm to 700~nm, and downsample the spectral resolution at intervals of 10~nm.  

To enable comprehensive spectral recovery across both the visible and near-infrared ranges, our proposed dataset retains the full-spectrum data while also providing visible-range HSIs with 31 channels. Additionally, our dataset contains significantly more images compared to other HSI datasets, making it uniquely comprehensive. Importantly, only our proposed BJTU-UVA dataset is suitable for auto-calibration, as other datasets do not include raw data essential for this task.  

Although this paper focuses on auto-calibration, the BJTU-UVA dataset is versatile and can support other tasks as well. The dataset will be publicly released in the future to facilitate further research.


\noindent
\textbf{Supplementary Reference}

\noindent
[35] Fumihito Yasuma, Tomoo Mitsunaga, Daisuke Iso, and Shree K. Nayar. Generalized assorted pixel camera: postcapture control of resolution, dynamic range, and spectrum. IEEE Transactions on Image Processing, 19(9):2241–2253, 2010.

\noindent
[36] Ayan Chakrabarti and Todd Zickler. Statistics of real-world hyperspectral images. In Proceedings of the IEEE/CVF Conference on Computer Vision and Pattern Recognition, pages 193–200, IEEE, 2011.

\noindent
[37] Rang MH Nguyen, Dilip K. Prasad, and Michael S. Brown. Training-based spectral reconstruction from a single RGB image. In European Conference on Computer Vision, pages 186–201. Springer, 2014.

\noindent
[38] Boaz Arad and Ohad Ben-Shahar. Sparse recovery of hyperspectral signal from natural RGB images. In European Conference on Computer Vision, pages 19–34. Springer, 2016.

\noindent
[39] Boaz Arad, Ohad Ben-Shahar, and Radu Timofte. NTIRE 2018 challenge on spectral reconstruction from RGB images. In Proceedings of the IEEE Conference on Computer Vision and Pattern Recognition Workshops, pages 929–938, 2018.

\noindent
[40] Boaz Arad, Radu Timofte, Ohad Ben-Shahar, Yi-Tun Lin, and Graham D. Finlayson. NTIRE 2020 challenge on spectral reconstruction from an RGB image. In Proceedings of the IEEE/CVF Conference on Computer Vision and Pattern Recognition Workshops, pages 446–447, 2020.

\noindent
[41] Boaz Arad, Radu Timofte, Rony Yahel, Nimrod Morag, Amir Bernat, Yuanhao Cai, Jing Lin, Zudi Lin, Haoqian Wang, Yulun Zhang, et al. NTIRE 2022 spectral recovery challenge and dataset. In Proceedings of the IEEE/CVF Conference on Computer Vision and Pattern Recognition, pages 863–881, 2022.

\noindent
[42] Du Z., Wei S., Liu T., et al. Exploring the applicability of spectral recovery in semantic segmentation of RGB images. IEEE Transactions on Multimedia, 2023.

\noindent
[43] Zhang M., Zhang C., Zhang Q., et al. ESSAformer: Efficient transformer for hyperspectral image super-resolution. In Proceedings of the IEEE/CVF International Conference on Computer Vision, pages 23073–23084, 2023.

\end{document}